\documentclass[letterpaper, 10 pt, conference]{ieeeconf}  % Comment this line out if you need a4paper
\usepackage{caption}
\usepackage{subcaption}

\usepackage{algorithm}
\usepackage{algorithmic}
\usepackage{mathtools}
\usepackage{balance}
\usepackage{dirtytalk}
\usepackage{svg}
\usepackage{hyperref}

\usepackage[algo2e,linesnumbered,ruled,vlined]{algorithm2e}

\usepackage{url}

\usepackage{todonotes}
\usepackage{amsmath}
\usepackage{amssymb}

\IEEEoverridecommandlockouts                              % This command is only needed if 
\DeclareCaptionLabelSeparator{periodspace}{.\quad}
\title{\LARGE \bf
Towards Confidence-guided Shape Completion for Robotic Applications
}

\author{Andrea Rosasco$^{1}$, Stefano Berti$^{1}$, Fabrizio Bottarel$^{1}$, Michele Colledanchise$^{1}$, and Lorenzo Natale$^{1}$ % <-this % stops a space
\thanks{$^{1}$The authors are with Istituto Italiano di Tecnologia, via San Quirico 19D, Genova, Italy ({\tt name.surname@iit.it})}
\thanks{This work was supported by the Italian National Institute for Insurance against Accidents at Work (INAIL) ergoCub Project.}
}

\begin{document}

\maketitle
\thispagestyle{empty}
\pagestyle{empty}

%%%%%%%%%%%%%%%%%%%%%%%%%%%%%%%%%%%%%%%%%%%%%%%%%%%%%%%%%%%%%%%%%%%%%%%%%%%%%%%%
\begin{abstract}
Many robotic tasks involving some form of 3D visual perception greatly benefit from a complete knowledge of the working environment. However, robots often have to tackle unstructured environments and their onboard visual sensors can only provide incomplete information due to limited workspaces, clutter or object self-occlusion.
% Robots often require 3D information about the objects they manipulate.
% Yet, their onboard sensors provide incomplete representation as objects view suffer, in the best case, from self-occlusion.
In recent years, deep learning architectures for shape completion have begun taking traction as effective means of inferring a complete 3D object representation from partial visual data. 
% A cost-effective and well-adopted solution employs a neural network to infer a complete representation of an object from incomplete data.
Nevertheless, most of the existing state-of-the-art approaches provide a fixed output resolution in the form of voxel grids, strictly related to the size of the neural network output stage. While this is enough for some tasks, e.g. obstacle avoidance in navigation, grasping and manipulation require finer resolutions and simply scaling up the neural network outputs is computationally expensive. In this paper, we address this limitation by proposing an object shape completion method based on an implicit 3D representation providing a confidence value for each reconstructed point. As a second contribution, we propose a gradient-based method for efficiently sampling such implicit function at an arbitrary resolution, tunable at inference time.

% we propose a novel method to sample from such an implicit function by updating random points with steps towards the gradient of the loss between the output of the implicit function and the value one.

% We validate our method with two sets of experiments, one comparing the reconstructed shapes with a ground truth; and one using the reconstructed shapes to grasp objects, using an existing grasping pipeline. In both sets of experiments, we compare the results with the state-of-the-art and provide guidelines to reproduce the experiments.

We experimentally validate our approach by comparing reconstructed shape with ground truth, and by deploying our shape completion algorithm in a robotic grasping pipeline. In both cases, we compare results with a state-of-the-art shape completion approach.
% We experimentally validate our proposed approach in two stages. First, we compare reconstructed shapes with ground truth using a popular object dataset. Second, we integrate our shape completion method in a state-of-the-art visual-based grasp planning pipeline. In both experiments, we compare results with a state-of-the-art shape completion deep learning architecture and provide guidelines to reproduce our experiments. 
The code is available at \url{https://github.com/andrearosasco/hyperpcr}

\end{abstract}

%%%%%%%%%%%%%%%%%%%%%%%%%%%%%%%%%%%%%%%%%%%%%%%%%%%%%%%%%%%%%%%%%%%%%%%%%%%%%%%%
\section{Introduction}
% The majority of high-level tasks that people would expect robots to perform involves grasping. While humans can easily grasp previously unseen objects the robots' ability to do that lag far behind.
% Point cloud extraction through depth estimation represent the most affordable and accurate technology to gather 3D information of a scene \cite{}.
% Applications span from 3D object tracking \cite{XXX} to grasp synthesis \cite{XXX}.

%In particular, in robotics, the possibility of obtaining the point cloud of an object of interest is important for synthesizing reliable grasps of such object.
% Unfortunately, the point cloud extracted from a single point of view suffers, in the best case, from self-occlusion, resulting in a incomplete representation. While it is still possible to estimate the 6DoF pose and generate grasps from incomplete representation, the estimation often relies on pre-existing complete 3D models. Such assumption jeopardizes the algorithm's generalization capabilities.

Nowadays, 3D environment perception for robotics is as affordable and available as it ever was. In some contexts, the environment the robot is operating in is completely known by design, e.g. industrial work cells, or can be thoroughly explored, simplifying tasks such as navigation or object manipulation. 
In many contexts, however, robots cannot rely on full knowledge of the working environment due to a limited workspace, unreachable viewpoints, or clutter, and only partial 3D representations are available. 
In the case of object grasping and manipulation, for instance, a robot might only be allowed partial views of a target object with unknown geometry. In this scenario, pose detection and classical grasp planning techniques cannot be employed. 
While it may still be possible to grasp and manipulate the object using partial 3D data, in latest years methods aiming at shape reconstruction (or \emph{shape completion}) using deep learning approaches have proven to be a promising research direction. 
% Deep shape completion methods learn the parameters of a neural network by solving an optimization problem. Prior knowledge is extracted from a labeled dataset allowing the neural network to generalize to unseen examples.
% A solution to this problem consists in taking the partial representation of the object, often called  \emph{2.5D representation}, and infer a complete representation. This task is known as \emph{shape completion}.

%One of the major differences between learning-based shape completion methods lies in the format used to represent the input and output shapes.
One of the first successful approaches to shape completion consists in extending Convolutional Neural Networks (CNNs) to 3D shapes. While these methods manage to achieve interesting results, they are constrained to a voxelized representation of their input and output shape. This results in a trade-off with the network complexity scaling cubically with the resolution. After the introduction of input layers capable of processing unstructured set of points, various works started to operate directly on point clouds. These approaches have the benefits of relying on an input format both lightweight and easy to extract from the data provided by 3D sensors.
\begin{figure}[t!]
    \centering
\includegraphics[width=0.9\columnwidth]{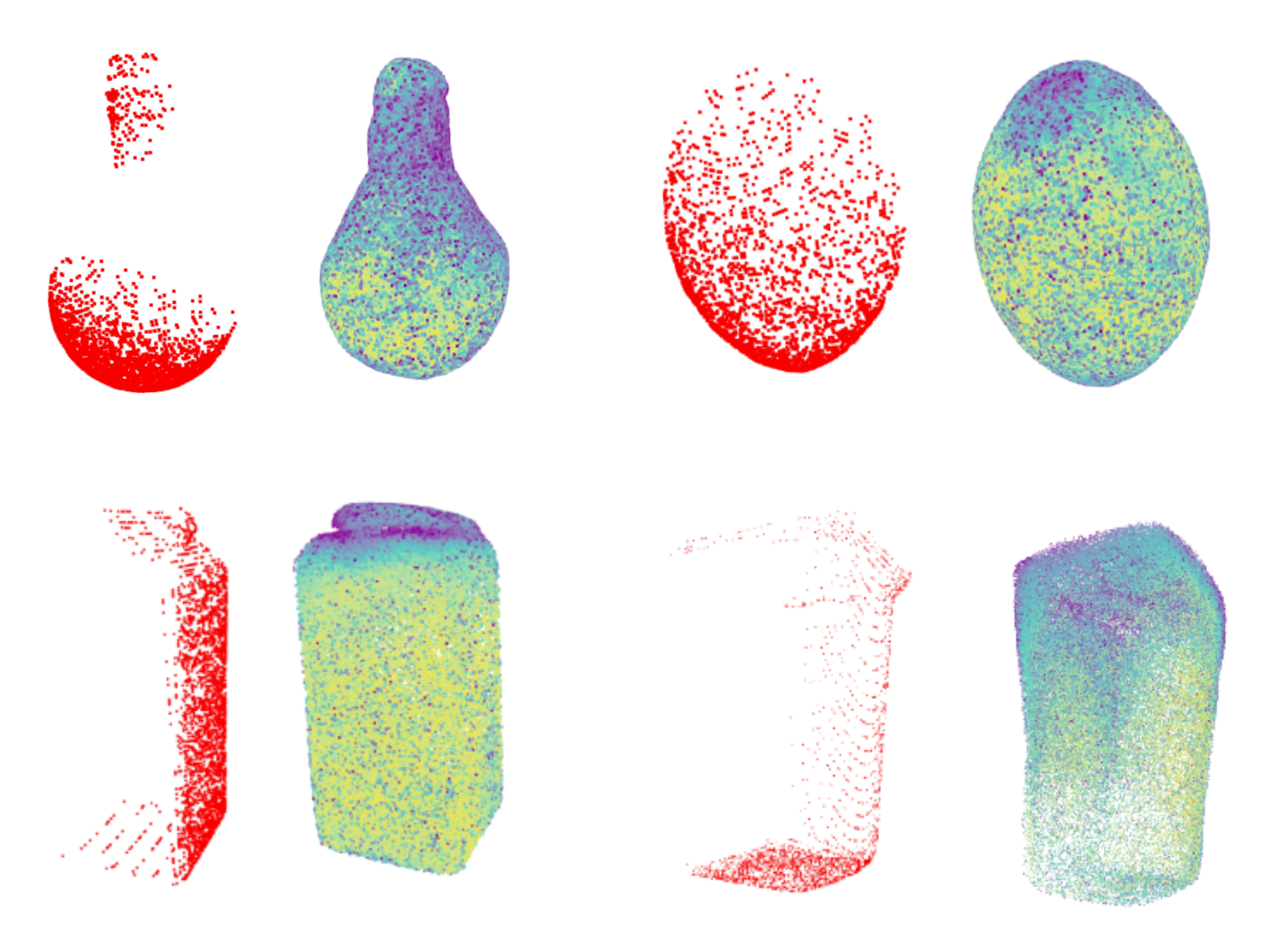}
    \caption{Partial point clouds and reconstructions of four objects. The color scheme of the reconstruction indicates the uncertainty of the model on each point (brighter means higher certainty). Particularly the model tends to assign higher probabilities to points near the visible part of the object. This could help in grasp evaluation.}
    \label{IN.fig.front}
\end{figure}
Despite the representative power of point clouds, methods that use them as output representation often suffer from fixed output resolution, as the number of generated points depends on the output layer dimension.
Learning an implicit representation yields different advantages such as reducing the number of parameters of the model and allowing for an arbitrary resolution of the output despite having a fixed-size implicit function.

To this end, we propose a framework that achieves arbitrary output resolution. It consists in generating an implicit function that captures a continuous model of the output space. We structured our model as a HyperNetwork \cite{Ha2016}, a particular neural network architecture where a primary network generates the weights for a secondary network used to perform the task. The structure allows to reduce the size of the model in terms of learnable parameters and implement a fine-tuning routine to better exploit the data available at evaluation time.

The main contributions of this work are:
\begin{itemize}
\item a novel hypernetwork-based  architecture for shape completion using a transformer decoder
\item a gradient-based sampling algorithm to reconstruct the point cloud from the implicit representation.
\end{itemize}

We experimentally validate our proposed approach in two stages. First, we compare reconstructed shapes with ground truth using a popular object dataset. Second, we integrate our shape completion method in a state-of-the-art visual-based grasp planning pipeline. In both experiments, we compare results with a state-of-the-art shape completion deep learning architecture and provide guidelines to reproduce our experiments. 

% We perform two experiments to validate our model. In the first one we assess the performance of our shape completion algorithm on a dataset of synthetic shapes using a popular metric. In the second one we use our approach to provide reconstructed point clouds to a grasp planning algorithm and perform grasping on real objects and we examine how shape completion affects the grasping performances.

The rest of the paper is organized as follows. Section \ref{section:related_work} briefly outlines works related to object modeling and shape completion. Section \ref{section:background} summarizes some background information useful in the understanding of this work. Section \ref{section:problem_formulation} formalizes the shape reconstruction problem and Section \ref{section:solution} details the proposed solution. Section \ref{section:experimental_evaluation} explains our experimental methodology and Section \ref{section:conclusion} closes off the paper by summarizing our results and proposing a future work direction.

\section{Related Work}
\label{section:related_work}

Existing shape completion algorithms can be classified based on the representation they use for their algorithm. Two popular shape representations are point clouds and voxels.

\subsection{Voxel-based}
Voxels are the extension of pixels to the three-dimensional space. While they can be represented with different data structures, the one commonly used for learning-based shape completion is that of an occupancy grid.
Occupancy grid representation allows a straightforward adaptation of convolutional neural networks to the 3D space \cite{choy20163d}. This approach has the advantage of both relying on a structured representation of the object and on a widely tested architecture (i.e., CNNs), but suffers from high computational cost. Since the number of operations scales cubically with the resolution of the occupancy grid in the current implementations, we cannot use them to process and generate shapes with fine-grained details.

Although some voxel-based algorithms can manage resolutions up to $128^{3}$ \cite{wu2018learning}, they require a considerable amount of memory constraining the algorithm to shallow architectures.

% \subsection{Mesh-based}
% Various approaches use meshes as a shape representation for the shape completion tasks. Most of them focus on using a parametrization technique to morph a 2D plane into a 3D surface. Since the parametrization algorithm is often sensitive to the input mesh quality, recent algorithms have been using deep neural networks to learn such parameterization.
% However, mesh-based solutions often suffer from self-intersecting meshes or cannot guarantee closed surfaces.
% Sphere parameterization may be used to generate a closed mesh, but the resulting shape is limited to the topological sphere.

\subsection{Point-based}
Point clouds are often used as a representation for 3D data, given their small memory footprint. Their sparse nature, however, precludes direct applicability of convolution operators.
% Given their small memory consumption, point clouds are often used as a representation for 3D objects. However, their unstructured nature precludes the direct application of convolutions.
PointNet \cite{qi2017pointnet} was one of the first methods to work directly on point clouds. Then, an increasing number of works started to use them as input and output representations. These methods usually feature a fixed size output layer, hence impeding the extraction of a denser point cloud. Moreover, since point clouds are often obtained through regressions, these algorithms are often unable to associate their output points to a confidence value.
%However, a primary limitation of learning with point clouds is that after training the output size is usually fixed and the output space cannot be sampled to gather a denser point cloud. 
% Option 2
\subsection{Shape Completion for Grasping}
Grasping algorithms can be categorized into model-based and model-free, depending on whether or not specific knowledge about the object (e.g., CAD model or previously scanned model \cite{bormann2019towards}) is required to solve the considered task.
Model-based algorithms \cite{tremblay2018deep} usually include a pose estimation step to match the model of the object to the real object point cloud. While this usually leads to precise grasps, it requires the object to be known a priori.
This assumption is oftentimes unrealistic.

Model-free approaches \cite{mahler2016dex} instead, do not need prior knowledge of the input object and operate just on its partial representation obtained from the robot sensors. While some grasping algorithms can directly operate on partial views, shape completion can help to increase the probability of finding a good grasp.

Some of the first algorithms that used completion were guided by the geometry of the object and attempted to reconstruct the occluded part exploiting symmetry \cite{bohg2011mind} or by the use of heuristics \cite{schnabel2009completion}. The work of Varley \cite{varley2017shape} was one of the first to pioneer deep learning-based shape completion applied to robotic grasping. Following their work \cite{lundell2019robust} developed a voxel-based architecture that uses Monte Carlo dropout to measure the uncertainty of the shape. More recent work uses implicit functions as an output representation in the form of signed distance functions \cite{yang2021robotic} or mapping functions \cite{yang2018foldingnet}.

\section{Background}
\label{section:background}

\subsection{HyperNetwork}
Hypernetwork \cite{Ha2016} refers to a particular neural network architecture composed of a primary network $f$ and a secondary network $g$, where $f$ is responsible for generating the weights of $g$ and $g$ generates the final output. This network architecture results particularly efficient in terms of parameters as the only learnable weights are the ones of the primary networks and it also directly learns an implicit function of the output space.
Thanks to their properties, Hyper-Networks were applied to image-to-mage translation \cite{jia2016dynamic}, neural architecture search (NAS) \cite{zhang2018graph} and shape reconstruction \cite{littwin2019deep}.

\subsection{Transformer}
Transformers \cite{vaswani2017attention} are an attention-based architecture first introduced as a solution for Natural Language Process (NLP) tasks. The fundamental mechanism of the transformer is self-attention. Attention consists in measuring the similarity of a query embedding against a set of key embedding through dot product. This produces a list of similarity scores that is used to take a weighted sum of the value embeddings. When this procedure is applied for each element of a sequence against all the other elements, it takes the name of self-attention.
Transformers are also based on an encoder-decoder structure that makes them perfect for sequence-to-sequence tasks (e.g. machine translation). Despite that, both the attention-based encoder and decoder of the transformer have been used by themselves with great success \cite{devlin2018bert}.
While the ability to pick up on long sequences of elements automatically made the transformer a good fit for language tasks, it was not immediately clear how to exploit such architectures in other tasks. The publication of the Vision Transformer (ViT) \cite{dosovitskiy2020image} showed how the architecture could attain good results in image recognition if properly designed. After ViT, an increasing number of publications \cite{liu2021swin} successfully applied transformers to computer vision tasks achieving state-of-the-art results.

\subsection{Geometry-aware Transformer}
Recently, after their first successful application to computer vision, the transformers have been successfully applied to shape reconstruction. PoinTr \cite{yu2021pointr} is a transformer-based architecture that uses a geometric-aware attention layer to introduce an inductive bias in the model.

Since the transformer computational complexity scales quadratically with the input sequence length, the point cloud cannot be directly passed as input to the transformer but is pre-processed to generate a shorter sequence of embedding $\mathcal{F} = \{\mathcal{F}_0, \,...,\, \mathcal{F}_N\}$ representing the point cloud.
Attention between the inputs' embeddings is computed using the standard self-attention equation:
\begin{equation} \label{eq:imp_func1}
        softmax(\frac{Q K^T}{\sqrt{d_k}})V.
\end{equation}
At the same time, following the work of \cite{yu2021pointr}, the edge between each proxy point and the $n$ closest proxy points are passed through a linear layer generating the edge embeddings:
\begin{equation} \label{eq:imp_func2}
        E_{i,\,j} = ReLU(\theta \cdot (\mathcal{F}_j - \mathcal{F}_i) + \phi \cdot \mathcal{F}_i).
\end{equation}
Then the $n$ edge embedding gets aggregated through max-pooling. The newly computed geometric embeddings have the same dimensionality as the point proxies, but they encode more information about their surroundings. Then, the output of the self-attention layer is concatenated to one of the geometric layers and mapped to its original dimension to generate the output.

Since PoinTr treats the shape completion task as a sequence-to-sequence problem on point proxies, the output of the transformer is a sequence of point proxies $\mathcal{H} = \{\mathcal{H}_0, \,...,\, \mathcal{H}_M\}$ describing the full shape. 
To reduce the computational complexity, $M$ has to be low. Therefore, to generate a fine-grained point cloud, a multi-scale approach is applied. 
In particular, similarly to PCN\cite{yuan2018pcn}, the point proxies are passed to a FoldingNet \cite{yang2018foldingnet} that deforms a 2D grid of points to generate the complete point cloud.
\subsection{Grasp Pose Detection (GPD)}
GPD \cite{ten2017grasp} is a deep learning method taking point clouds as input in order to produce 6D grasp candidates with no a priori knowledge of the target CAD model. It works by evenly sampling candidates on the point cloud surface and filtering them according to geometric criteria. Each raw candidate is processed through a CNN to obtain a measure of quality, indicating how much each candidate resembles a frictionless antipodal grasp. To convert such input into a CNN-compatible representation, the portion of the point cloud enclosed in the gripper volume is voxelized. Three images from the perspective of each axis of the hand reference frame are then computed: the heightmap of the occupied voxels, the heightmap of the unoccupied voxels and the surface normal map. This results in a 9-channel image that can be processed by a CNN.
% Akin to object detection, GPD works in two steps: first, it evenly samples several grasps candidates on the object surface. Then a neural network takes each candidate and determine the grasp feasibility. The second step, called grasp classification, is executed through a convolutional neural network trained to predict if closing the gripper from the given input pose would result in a frictionless antipodal grasp. To convert the input into a representation suitable from a CNN, the part of the point cloud lying inside the gripper closing area is voxelized. Then we compute three images from the perspective of each axis of the hand reference frame: the heightmap of the occupied voxels, the heightmap of the unoccupied voxels and the surface normal map. This results in a 9-channel image that can be processed by a CNN.

\section{Problem Formulation}
\label{section:problem_formulation}
    Given an object $O$, let $Y \triangleq PC_{3D}(O) \in \mathbb{R}^{n \times 3}$ be a complete point cloud of  $O$, and $X \triangleq  PC_{2.5D}(O)\in \mathbb{R}^{n \times 3}$ a partial self-occluded point cloud. Let $m : \mathbb{R}^{n \times 3}, \mathbb{R}^{n \times 3} \rightarrow \mathbb{R}$ be a metric that gives a measure of the dissimilarity between the two point clouds and $\delta$ be a confidence bound. We define the shape completion problem as the problem of finding function $f :  \mathbb{R}^{n \times 3} \to  \mathbb{R}^{n \times 3}$ such that  $f(X) = \widetilde{Y} $  and $m(Y, \widetilde{Y}) \leq \delta$

\begin{figure*}[h]
    \centering
    \includegraphics[width=1\textwidth]{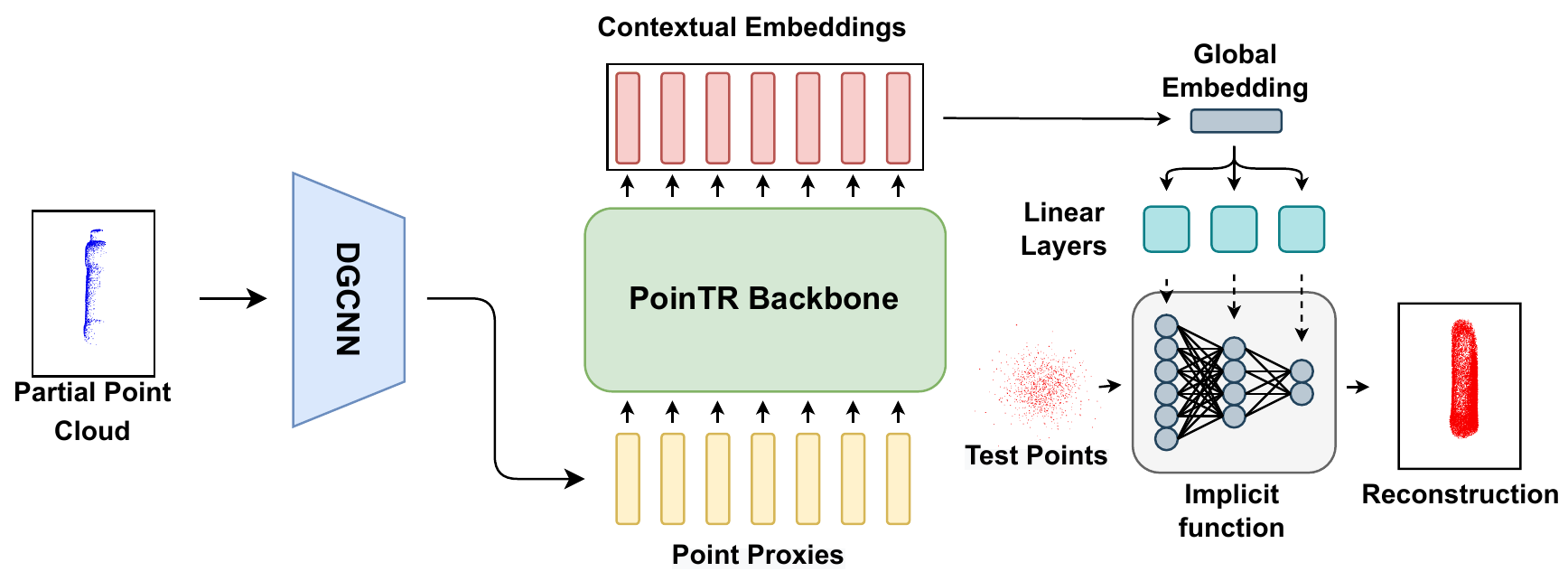}
    \caption{Overview of our shape completion architecture. The backbone computes the contextual embeddings of the input point proxies. Then, we pass the embeddings through a  max-pooling layer to obtain the global embedding. Finally, we feed the global embedding to different linear layers to generate weight, biases and scales of the implicit function.}
    \label{Figure: tiled}
\end{figure*}

\section{Proposed Solution}
\label{section:solution}
In this section we introduce our architecture to estimate a 3D reconstruction of an object from its 2.5D point cloud, in the form of an implicit function of the 3D space. To achieve arbitrary resolution and obtain a measure of uncertainty of each output point, we define an implicit shape representation that can be sampled on real coordinates  whose values gives us information about the complete shape of the object.
We adopt an hyper-network structure where the weights $\phi$ of an implicit function $g_{\phi}$ describing the complete shape of the object are generated by the backbone $f$.
Using this type of architecture, we manage to reduce the number of parameters of the model while obtaining good performances on the shape completion task.
We define $g_{\phi}$ as a parametric function
\begin{equation} \label{eq:imp_func3}
        g_{\phi}: \mathbb{R}^3 \rightarrow \{0,\, 1\}
\end{equation}
such that
\begin{equation} \label{eq:imp_func4}
            g_\phi(p)= 
\begin{cases}
    1,& \text{if } p \in Y\\
    0,              & \text{otherwise}
\end{cases}
\end{equation}

where $p=(x, y, z)$ is a tridimensional point.

To condition $g_{\phi}$ on past knowledge extracted from a dataset, we introduce an encoder module $f_\theta$ that takes in input a partial point cloud $X \in \mathbb{R}^{n \times 3}$ and returns the weights $\phi$ that parameterize the implicit function $g_{\phi}$.

\subsection{Implicit Function}
We implemented the implicit function $g$ as a multi-layer perceptron whose weights are generated by a backbone every time a partial input is processed.

Following previous work \cite{littwin2019deep} we generate three sets of weights for each layer $l$:
$$\phi_{l} = \{\phi_{l}^{W}, \phi_{l}^{b}, \phi_{l}^{s}\}$$
respectively the weight, bias and scale parameters.
Given a specific layer $l$, its input $x_l$, and the parameters $\phi_{l}$, the output of the layer is computed as follows:
\begin{equation} \label{eq:layer_comp}
        y_{l} = (( \phi_{l}^{W} \, x_{l}) \cdot \phi_{l}^{s}) + \phi_{l}^{b}.
\end{equation}
In particular, we use a multi-layer perception with two hidden layers of dimension 32 with Leaky ReLU activation functions set to 0.2.

\subsection{Backbone}
As the backbone of our model, we use the decoder of the PoinTR transformer introduced in Section \ref{section:background} with one geometric layer in the first transformer block. Following their example, we compute the point proxies of the input point cloud using a hierarchical version of Dynamic Graph CNN \cite{wang2019dynamic}. This input layer uses EdgeConv operator to reduce the full point cloud to a fixed number of embeddings that we call point proxies, each representing a different part of the input. 
After computing the $N$ point proxies, we input them to the transformer as a sequence and get a contextual embedding for each of them. Thanks to self-attention, these embeddings represents each point proxy in relation to the other ones.
To get a global representation of the point cloud, we apply max pooling to the embeddings and obtain the global embedding $e$.
The global embedding is then processed by a set of linear layers to generate the weights $\phi$ for each layer of the implicit function.

\subsection{Implicit Function Input}
During training, in addition to the partial point cloud taken as input by the backbone, we create a batch of points spanning the output space to train the implicit function. We sample half of the points from the ground truth point cloud and label them as positive examples. The other half is composed by a 10\% of points uniformly sampled across the output space and a 40\% obtained by adding Gaussian noise to points from the ground truth. This is motivated by the fact that points sampled near the objects' boundaries holds more information about the shape of the object and facilitate training.
To assign a label to the perturbed and the uniformly sampled points, we check their distance to the points on the ground truth. If they are close enough to it they are assigned a positive label.

We then feed the batch of points to the implicit function that returns, for each point, a value between $0$ and $1$ using a sigmoid activation function. We compute the binary cross entropy loss between the predictions and the labels and backpropagate the gradient to the backbone that updates its parameters. This results in learning a model that is able to map a partial point cloud to a function representing its respective complete point cloud.

\subsection{Point Cloud Sampling}
Once the backbone has produced an implicit function representation of the complete shape, we might need to sample a point cloud from it. The naive approach would be to create a grid of points and test each of them through the point cloud. The issue with this method is that just the small fraction of points that intersect the implicit surface will contribute to forming the point cloud. This problem worsens if we increase the classification threshold to gather points with a higher probability of being part of the object. Furthermore, increasing the sampling resolution increases the number of points to be tested cubically, increasing the risk of running out of memory. To solve this problem, we developed an algorithm that uses the first-order information associated with each point to iteratively reconstruct the input point cloud. Our algorithm steps are shown in Algorithm\ref{algorithm: pc-sampling}. Note that in our implementation we omit the for loop in Line \ref{algline: pointloop} as we operate on matrices of points.
We initialize a list of point $\mathcal{L}$ by randomly sampling from a Gaussian distribution centred in $0$ and with standard deviation $0.1$ (Line 2). Then, for each point, we get the output of the implicit function and add it to the output if its probability is higher than the threshold (Lines 4 to 9).
After that, we measure the loss between the probability assigned from the implicit function and probability $1$. Through back-propagation we compute the gradient of the loss function with respect to the input point and update it through gradient descent (Line 10). Finally, we update the point in $\mathcal{L}$.
In short, the points in $\mathcal{L}$ move toward the surface described by the implicit function $g_\phi$ and every time one of the points crosses the threshold we copy it and add it to our output point cloud.

\begin{center}
\begin{algorithm}
\small
\caption{Point Cloud Sampling} \label{algorithm: pc-sampling}
\textbf{Inputs:} $g_\phi$: implicit function; $N$: number of points to be sampled \\
\textbf{Inputs:} $\eta$: step size; $I$: number of steps; $\tau$: threshold; $\ell$: loss function\\
\begin{algorithmic}[1]
\STATE $reconstruction\leftarrow$empty list
\label{algorithm: line2}
\STATE $points\leftarrow$list of N 3D points sampled from $\mathcal{N}(0, 0.1)$
\WHILE{$output.Size() < N$}
    \FOR{$i=1 \,\,\,\boldsymbol{to}\,\,\, points.Size()$} \label{algline: pointloop}
    \STATE $x=points[i]$
    \STATE $\tilde{y} = g_\phi(x)$
    \IF{$\tilde{y} > \tau$}
        \STATE $output.insert(x)$
    \ENDIF
    \STATE $points[i] = x - \eta \nabla_{x} \ell (\tilde{y}, 1)$
    \ENDFOR
\ENDWHILE
\end{algorithmic}
\textbf{Output:} $reconstruction$
\end{algorithm}

\end{center}

\section{Experimental Validation}
\label{section:experimental_evaluation}
We run two sets of experiments: one where we evaluate the performance of our shape completion algorithm and compare it with the state of the art; and one where we give the reconstructed shape to a grasping algorithm and measure the grasp success rate. We compare our shape completion results with the one presented by Varley \cite{varley2017shape} (V) and the one presented by Lundell \cite{lundell2019robust} (USN).
Then we choose GPD as our grasp planner and compare our grasping pipeline with one composed of USN shape completion and GPD.
Both V and USN use a 3D CNN to generate a $40\times40\times40$ voxel grid of the complete shape. Then, to convert their output to a representation more suitable for grasping, they apply a surface reconstruction algorithm introduced in \cite{varley2017shape} to generate a mesh. The CNN used by Varley applies three convolutions to the input voxel grid followed by two dense layers. The output of the dense layer is then reshaped in the final voxel grid. USN instead adopt an encoder-decoder architecture with skip connections that reconstruct the final shape through deconvolution.  Furthermore, USN applies dropout at inference time to generate different reconstructions of the same input. The reconstructions are then averaged to generate the complete shape.

\subsection{Dataset}
We trained our model on self-occluded point clouds generated from meshes from the Columbia Grasp Database \cite{goldfeder2009columbia} and YCB \cite{calli2015ycb}. Following the procedure used in \cite{lundell2019robust} we split the test data in
\begin{itemize}
    \item Holdout Views: views of objects that have already been observed during training, but from different angles.
    \item Holdout models: views of objects never observed during training.
\end{itemize}
We use the same splits provided by V. This resulted in about $200$k training samples, $60$k holdout views and $100$k holdout models. We also created a small validation set containing all the views of $4$ models from the training set to be used for early stopping and hyper-parameter selection.

The partial views of the objects were generated by loading the models in Gazebo and rendering their depth image. Each object was rotated around the three axes to generate a total of 726 partial views.

\begin{figure}[h]
    \centering
    \includegraphics[width=0.48\textwidth]{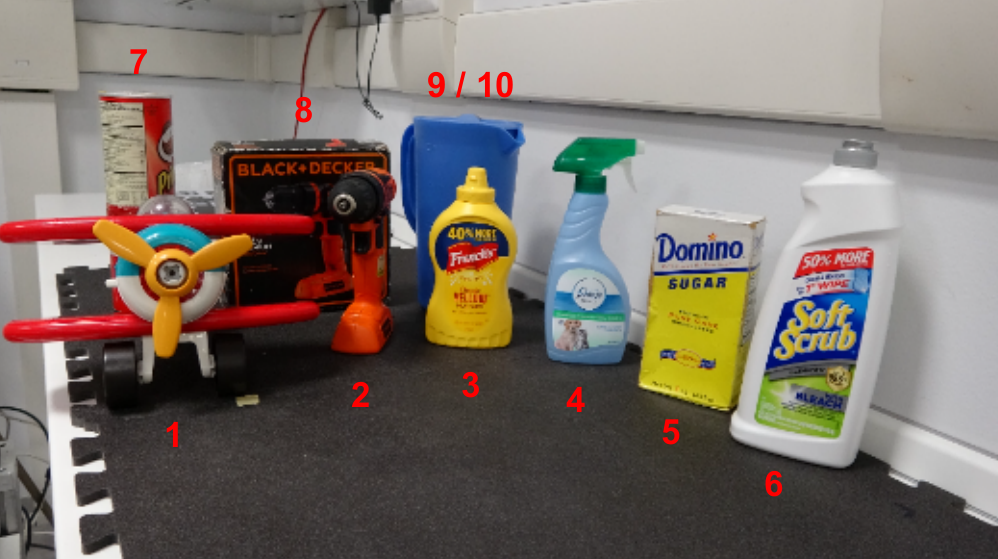}
    \caption{Official YCB names following the numbering order and skipping the object number 8 (not official YCB object): airplane toy, power drill, mustard container, glass cleaner, box of sugar, bleach cleanser, chips can, pitcher. We consider object number $10$ as the pitcher without its lid on.}
    \label{Figure: objects}
\end{figure}

\subsection{Training Procedure}
We trained our model on an NVIDIA V100 for 60 epochs. The model was updated with Adam optimizer with a learning rate of $1e^{-4}$ and a mini-batch size of $32$. We set the encoder depth to $4$. We use $6$ attention heads and an input embedding dimension of $384$. For the grasping experiments, we perturbed the input point cloud with noise to make it more robust to the input provided by the depth camera.

\subsection{Shape Completion}

\begin{table}[h]
\centering
\begin{tabular}{|l|l|l|l|}

\hline
               & V & USN    & Ours \\ \hline
Training Views & 0.6205 & 0.6446 & \textbf{0.6712}                    \\ \hline
Holdout Views  & 0.6143 & 0.6389 & \textbf{0.6667}              \\ \hline
Holdout Models & 0.5632 & 0.5651 & \textbf{0.6023}              \\ \hline
%               & USN    & Ours   & Ours (full dataset) \\ \hline
% Training Views & 0.6446 &        & 0.6712                    \\ \hline
% Holdout Views  & 0.6389 & 0.6331 & 0.6667              \\ \hline
% Holdout Models & 0.5651 & 0.6116 & 0.6023              \\ \hline
\end{tabular}
\caption{Jaccard Similarity results. \cite{varley2017shape} and \cite{lundell2019robust} reported their result on a random sample of 50 examples per split. For our algorithm we report the results on all the samples in each of the three splits.}
\label{tab:jaccard}
\end{table}
To evaluate their algorithms, Varley and Lundell, generated the test data by sampling 50 views from the training set, 50 views from the holdout views set and 50 views from the holdout models set. We assume that they did not test on the full dataset due to the meshing process being too computationally expensive. Unfortunately, we did not manage to use the surface reconstruction algorithm introduced by V and used by USN to replicate their results. For this reason, we report the results as indicated in their work. Testing only our architecture on randomly sampled subsets of 50 of the three splits would not provide a good comparison: two different subset of dimension 50 from a set of $100$k elements can lead to very different result. For this reason, we report the results of our algorithm computed on the whole splits.

We assess the shape completion performance of our algorithm against USN by measuring the Jaccard Similarity between the ground truth and our reconstruction:
\begin{equation}
    J(A, B) = \frac{|A\cap B|}{|A\cup B|}
\end{equation}
In particular, we first convert our prediction and the corresponding ground truth to a voxel grid of size $40^3$ and then check the number of voxel occupied by their union and by their intersection. To generate our output point cloud we used the sampling technique showed in Section \ref{section:solution}. We set the generated number of points to $1e^5$, then we downsample the point cloud with farthest point sampling to $16384$ to obtain a more uniform reconstruction.

The average results of our method and USN are showed in Table \ref{tab:jaccard}. We can see that our algorithm is able to score a higher Jaccard similarity on the three splits.
To attain our results we selected on the validation set the best parameter for our sampling algorithm: $20$ iterations, a threshold of $0.85$ and an update step of $0.1$.

\subsection{Grasping Experiments}
\begin{figure}[h!]
    \centering
\includegraphics[width=0.9\columnwidth]{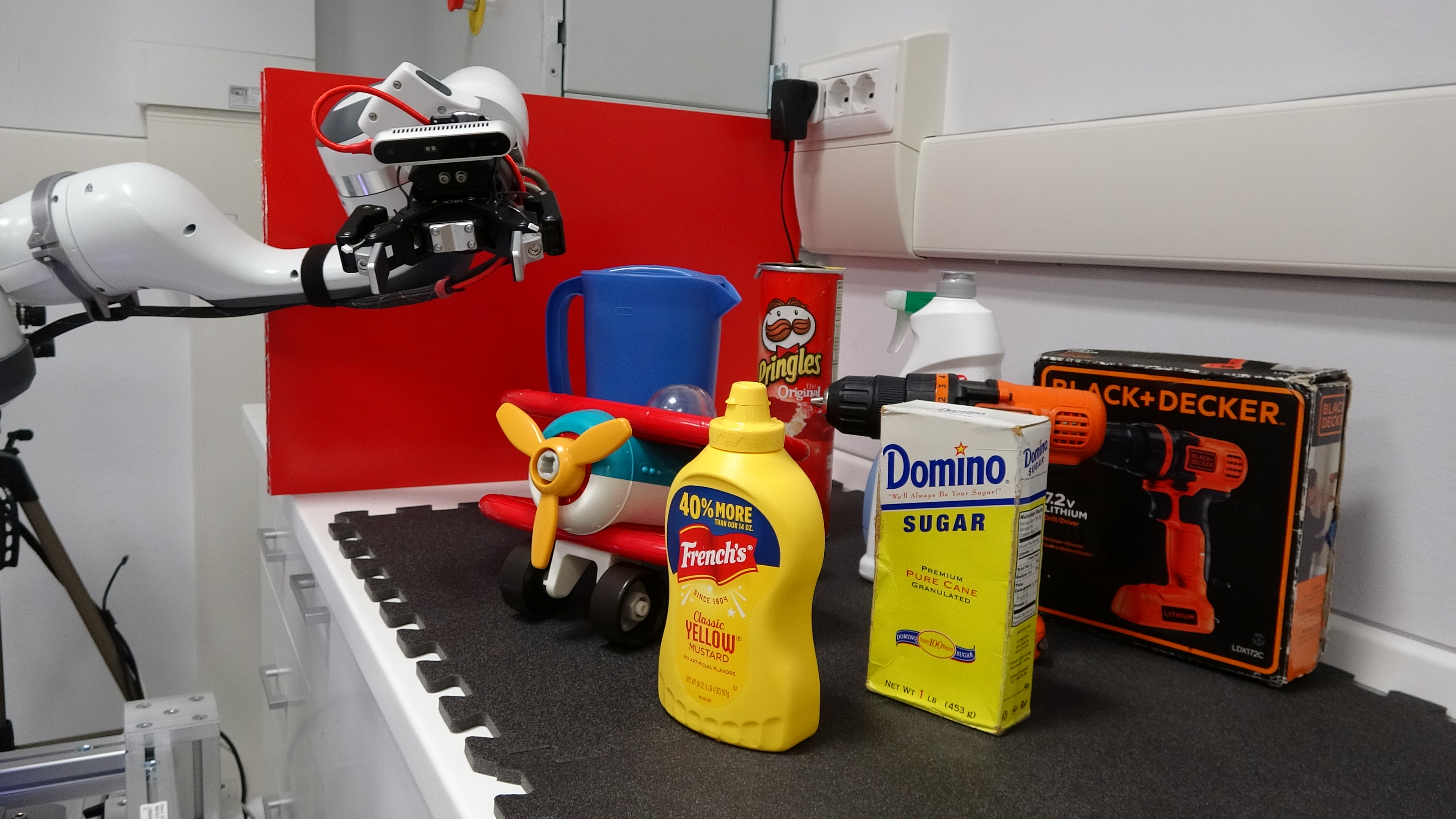}
    \caption{Our robotic grasping set-up.}
    \label{Figure: grasping-set-up}
\end{figure}

Like \cite{lundell2019robust}, for the grasping experiments we used 8 holdout YCB objects plus the power drill box. The objects are shown in figure \ref{Figure: objects}. We also tried to grasp the pitcher (number $9$) without its lid on and considered it object number $10$. The robotic platform consists in a Franka Emika Panda arm equipped with a Robotiq 2F-85 gripper and a RealSense D415. The setup is shown in Figure \ref{Figure: grasping-set-up}.
Following the same protocol as \cite{lundell2019robust}, we attempted a total of 10 grasps per object on five different orientations (0°, 72°, 144°, 216°, and 288°) for a total of 100 grasps per method. Differently from \cite{lundell2019robust}, our camera is hand-mounted and the 2F-85 gripper features a much smaller graspable volume with respect to the Barrett hand used in that work. In order to fairly compare the effectiveness of USN against our approach in a grasping scenario, we generate completed point clouds using both algorithms and feed them in turn to the GPD planner. A grasp is considered successful if the robot is able to lift the object off the table and hold it for $5$ seconds. 

% To account for these differences in the two set-ups and the use of a different grasping algorithm, we created a pipeline based on USN shape completion and GPD and compare it against our algorithm.
% We used marching cubes to convert their output voxel grid into a mesh. Then, we sample a point cloud from the mesh and pass it to GPD to compute the grasp. 

% We considered a grasp successful if the gripper could lift the object from the table and hold it until it was automatically released $5$ seconds later.

\begin{figure}[!h]
    \centering
    \includegraphics[width=.44\textwidth]{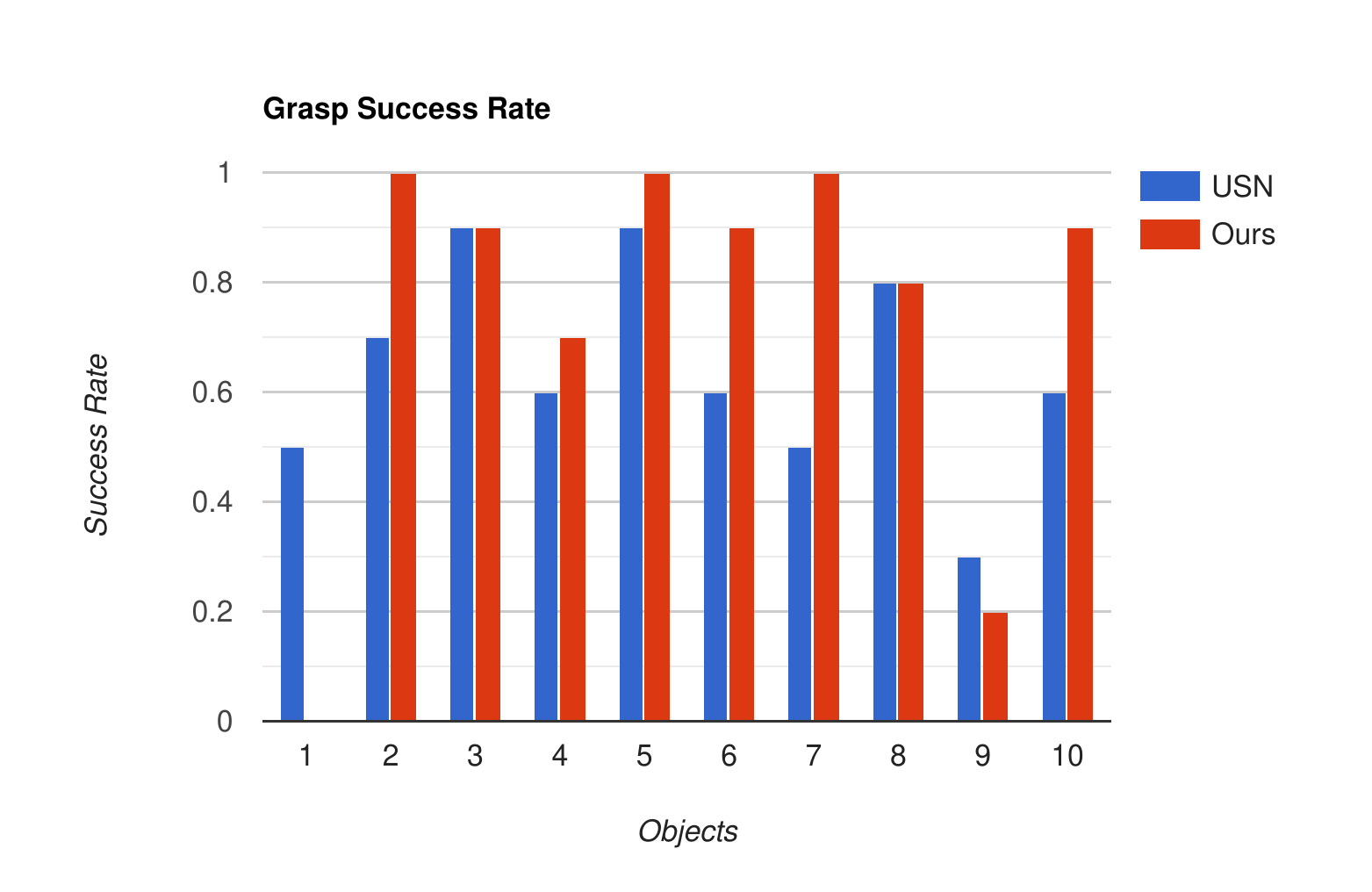}
    \caption{Comparison of the grasp success rate of our pipeline and USN pipeline on each object.}
    \label{Figure: bar-graph}
\end{figure}

The results on each test object are reported in Figure \ref{Figure: bar-graph}.
It is interesting to notice how, on average, the success rates achieved by USN in our experiments are higher than the results reported in their work. This is most probably due to the use of GPD as a grasping algorithm and its robustness to noisy inputs. 

\begin{figure}[!h]
    \centering
    \includegraphics[width=.38\textwidth]{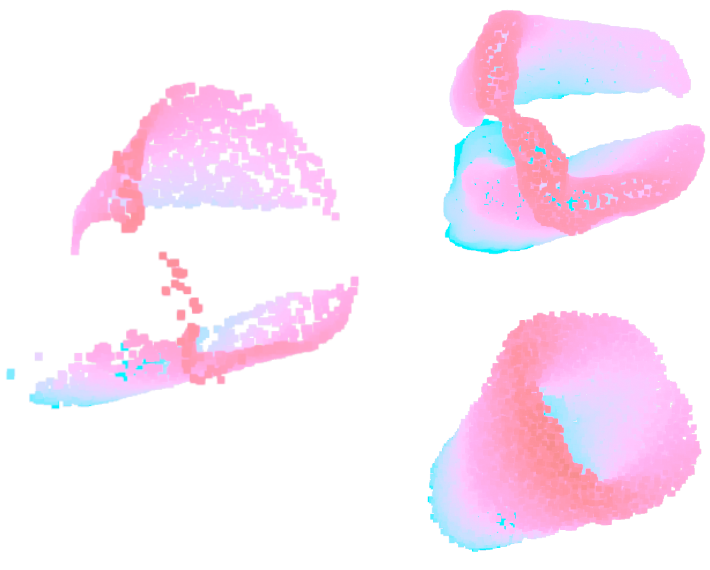}
    \caption{Left: partial point cloud of the pitcher without its lid. Top right: USN reconstruction. Bottom right: our reconstruction. Note the correspondence between the partial shape and USN reconstruction. The colors are just used to give a sense of perspective.}
    \label{Figure: pitcher}
\end{figure}

The only object on which USN performances are considerably lower than the one obtained in their experiments is the pitcher. The reason is that the gripper they used, a Barrett hand, is big enough to grasp the pitcher without using its handle. Our gripper is considerably smaller and could grasp the pitcher just from its handle.

While our method struggles with some of the more complex objects, on average we perform better than USN. In particular, we score a total 75\% success rate against USN 65\%.

We observed that this is due to USN's tendency to reconstruct the objects just by increasing the volume around the partial observation. It seems that the predictions generated by USN prioritize precision over recall, generating more voxels close to the partial observation. This means that several times USN reconstructions did not have enough depth, causing the predicted grasp to collide with the object. This can be noticed in Figure \ref{Figure: pitcher}.

Our algorithm on the other end tends to favour recall, reconstructing the surface of the objects without sticking too close to the partial observation. As shown in Figure \ref{IN.fig.front}, while the area closer to the input point cloud has a higher probability, the whole shapes are correctly reconstructed.

The grasping success rate on object $1$, the toy aeroplane, reflects the tendency of our algorithm to favour recall. Our method consistently filled the space between the upper and lower wing making it almost impossible to grasp. This is not surprising since with respect to the training data object $1$ is out of distribution.

\section{Conclusion}
\label{section:conclusion}
We presented a novel algorithm for shape completion based on an hypernetwork-based architecture that uses the decoder of a transformer to generate the weights of an occupancy function.
Having an implicit function as our output representation allows us to get a measure of confidence for each point of the output space.
Moreover, despite the size of the occupancy function being fixed, we can use it to sample a point cloud at arbitrary resolution.
We also introduced a technique to sample a dense point cloud of high confidence points from our implicit function, using the gradient to maximize the points' probability. 

Quantitative experiments demonstrate superior performances of our model for the shape completion task on the Columbia Grasp Database. We tested on real hardware how the quality of the reconstruction affects the grasping success rate, showing that our reconstructions lead to better performance with respect to the previous state-of-the-art.

In future work, we would like to explore how the confidence measure provided by our model can be used to influence the output of a grasping algorithm, for example favouring grasps near high confidence areas of the reconstruction.

% \section{Acknowledgment}
% The paper was supported by the Italian National Institute for Insurance against Accidents at Work (INAIL) ergoCub Project.

\bibliographystyle{unsrt} % We choose the "plain" reference style
\bibliography{sc} % Entries are in the refs.bib file

\begin{thebibliography}{10}

\bibitem{Ha2016}
David Ha, Andrew Dai, and Quoc~V. Le.
\newblock Hypernetworks.
\newblock {\em Hypernetworks in the Science of Complex Systems}, pages
  151--176, 9 2016.

\bibitem{choy20163d}
Christopher~B Choy, Danfei Xu, JunYoung Gwak, Kevin Chen, and Silvio Savarese.
\newblock 3d-r2n2: A unified approach for single and multi-view 3d object
  reconstruction.
\newblock In {\em European conference on computer vision}, pages 628--644.
  Springer, 2016.

\bibitem{wu2018learning}
Jiajun Wu, Chengkai Zhang, Xiuming Zhang, Zhoutong Zhang, William~T Freeman,
  and Joshua~B Tenenbaum.
\newblock Learning shape priors for single-view 3d completion and
  reconstruction.
\newblock In {\em Proceedings of the European Conference on Computer Vision
  (ECCV)}, pages 646--662, 2018.

\bibitem{qi2017pointnet}
Charles~R Qi, Hao Su, Kaichun Mo, and Leonidas~J Guibas.
\newblock Pointnet: Deep learning on point sets for 3d classification and
  segmentation.
\newblock In {\em Proceedings of the IEEE conference on computer vision and
  pattern recognition}, pages 652--660, 2017.

\bibitem{bormann2019towards}
Richard Bormann, Bruno Ferreira~de Brito, Jochen Lindermayr, Marco Omainska,
  and Mayank Patel.
\newblock Towards automated order picking robots for warehouses and retail.
\newblock In {\em International Conference on Computer Vision Systems}, pages
  185--198. Springer, 2019.

\bibitem{tremblay2018deep}
Jonathan Tremblay, Thang To, Balakumar Sundaralingam, Yu~Xiang, Dieter Fox, and
  Stan Birchfield.
\newblock Deep object pose estimation for semantic robotic grasping of
  household objects.
\newblock In {\em Conference on Robot Learning}, pages 306--316. PMLR, 2018.

\bibitem{mahler2016dex}
Jeffrey Mahler, Florian~T Pokorny, Brian Hou, Melrose Roderick, Michael Laskey,
  Mathieu Aubry, Kai Kohlhoff, Torsten Kr{\"o}ger, James Kuffner, and Ken
  Goldberg.
\newblock Dex-net 1.0: A cloud-based network of 3d objects for robust grasp
  planning using a multi-armed bandit model with correlated rewards.
\newblock In {\em 2016 IEEE international conference on robotics and automation
  (ICRA)}, pages 1957--1964. IEEE, 2016.

\bibitem{bohg2011mind}
Jeannette Bohg, Matthew Johnson-Roberson, Beatriz Le{\'o}n, Javier Felip, Xavi
  Gratal, Niklas Bergstr{\"o}m, Danica Kragic, and Antonio Morales.
\newblock Mind the gap-robotic grasping under incomplete observation.
\newblock In {\em 2011 IEEE international conference on robotics and
  automation}, pages 686--693. IEEE, 2011.

\bibitem{schnabel2009completion}
Ruwen Schnabel, Patrick Degener, and Reinhard Klein.
\newblock Completion and reconstruction with primitive shapes.
\newblock In {\em Computer Graphics Forum}, volume~28, pages 503--512. Wiley
  Online Library, 2009.

\bibitem{varley2017shape}
Jacob Varley, Chad DeChant, Adam Richardson, Joaqu{\'\i}n Ruales, and Peter
  Allen.
\newblock Shape completion enabled robotic grasping.
\newblock In {\em 2017 IEEE/RSJ international conference on intelligent robots
  and systems (IROS)}, pages 2442--2447. IEEE, 2017.

\bibitem{lundell2019robust}
Jens Lundell, Francesco Verdoja, and Ville Kyrki.
\newblock Robust grasp planning over uncertain shape completions.
\newblock In {\em 2019 IEEE/RSJ International Conference on Intelligent Robots
  and Systems (IROS)}, pages 1526--1532. IEEE, 2019.

\bibitem{yang2021robotic}
Daniel Yang, Tarik Tosun, Benjamin Eisner, Volkan Isler, and Daniel Lee.
\newblock Robotic grasping through combined image-based grasp proposal and 3d
  reconstruction.
\newblock In {\em 2021 IEEE International Conference on Robotics and Automation
  (ICRA)}, pages 6350--6356. IEEE, 2021.

\bibitem{yang2018foldingnet}
Yaoqing Yang, Chen Feng, Yiru Shen, and Dong Tian.
\newblock Foldingnet: Point cloud auto-encoder via deep grid deformation.
\newblock In {\em Proceedings of the IEEE conference on computer vision and
  pattern recognition}, pages 206--215, 2018.

\bibitem{jia2016dynamic}
Xu~Jia, Bert De~Brabandere, Tinne Tuytelaars, and Luc~V Gool.
\newblock Dynamic filter networks.
\newblock {\em Advances in neural information processing systems}, 29, 2016.

\bibitem{zhang2018graph}
Chris Zhang, Mengye Ren, and Raquel Urtasun.
\newblock Graph hypernetworks for neural architecture search.
\newblock {\em arXiv preprint arXiv:1810.05749}, 2018.

\bibitem{littwin2019deep}
Gidi Littwin and Lior Wolf.
\newblock Deep meta functionals for shape representation.
\newblock In {\em Proceedings of the IEEE/CVF International Conference on
  Computer Vision}, pages 1824--1833, 2019.

\bibitem{vaswani2017attention}
Ashish Vaswani, Noam Shazeer, Niki Parmar, Jakob Uszkoreit, Llion Jones,
  Aidan~N Gomez, {\L}ukasz Kaiser, and Illia Polosukhin.
\newblock Attention is all you need.
\newblock {\em Advances in neural information processing systems}, 30, 2017.

\bibitem{devlin2018bert}
Jacob Devlin, Ming-Wei Chang, Kenton Lee, and Kristina Toutanova.
\newblock Bert: Pre-training of deep bidirectional transformers for language
  understanding.
\newblock {\em arXiv preprint arXiv:1810.04805}, 2018.

\bibitem{dosovitskiy2020image}
Alexey Dosovitskiy, Lucas Beyer, Alexander Kolesnikov, Dirk Weissenborn,
  Xiaohua Zhai, Thomas Unterthiner, Mostafa Dehghani, Matthias Minderer, Georg
  Heigold, Sylvain Gelly, et~al.
\newblock An image is worth 16x16 words: Transformers for image recognition at
  scale.
\newblock {\em arXiv preprint arXiv:2010.11929}, 2020.

\bibitem{liu2021swin}
Ze~Liu, Yutong Lin, Yue Cao, Han Hu, Yixuan Wei, Zheng Zhang, Stephen Lin, and
  Baining Guo.
\newblock Swin transformer: Hierarchical vision transformer using shifted
  windows.
\newblock In {\em Proceedings of the IEEE/CVF International Conference on
  Computer Vision}, pages 10012--10022, 2021.

\bibitem{yu2021pointr}
Xumin Yu, Yongming Rao, Ziyi Wang, Zuyan Liu, Jiwen Lu, and Jie Zhou.
\newblock Pointr: Diverse point cloud completion with geometry-aware
  transformers.
\newblock In {\em Proceedings of the IEEE/CVF international conference on
  computer vision}, pages 12498--12507, 2021.

\bibitem{yuan2018pcn}
Wentao Yuan, Tejas Khot, David Held, Christoph Mertz, and Martial Hebert.
\newblock Pcn: Point completion network.
\newblock In {\em 2018 International Conference on 3D Vision (3DV)}, pages
  728--737. IEEE, 2018.

\bibitem{ten2017grasp}
Andreas ten Pas, Marcus Gualtieri, Kate Saenko, and Robert Platt.
\newblock Grasp pose detection in point clouds.
\newblock {\em The International Journal of Robotics Research},
  36(13-14):1455--1473, 2017.

\bibitem{wang2019dynamic}
Yue Wang, Yongbin Sun, Ziwei Liu, Sanjay~E Sarma, Michael~M Bronstein, and
  Justin~M Solomon.
\newblock Dynamic graph cnn for learning on point clouds.
\newblock {\em Acm Transactions On Graphics (tog)}, 38(5):1--12, 2019.

\bibitem{goldfeder2009columbia}
Corey Goldfeder, Matei Ciocarlie, Hao Dang, and Peter~K Allen.
\newblock The columbia grasp database.
\newblock In {\em 2009 IEEE international conference on robotics and
  automation}, pages 1710--1716. IEEE, 2009.

\bibitem{calli2015ycb}
Berk Calli, Arjun Singh, Aaron Walsman, Siddhartha Srinivasa, Pieter Abbeel,
  and Aaron~M Dollar.
\newblock The ycb object and model set: Towards common benchmarks for
  manipulation research.
\newblock In {\em 2015 international conference on advanced robotics (ICAR)},
  pages 510--517. IEEE, 2015.

\end{thebibliography}
\end{document}